# Unsupervised Online Learning With Multiple Postsynaptic Neurons Based on Spike-Timing-Dependent Plasticity Using a TFT-Type NOR Flash Memory Array

Soochang Lee, Chul-Heung Kim, Seongbin Oh, Byung-Gook Park, and Jong-Ho Lee[*]

*Department of Electrical and Computer Engineering and the Inter-University Semiconductor Research Center (ISRC), Seoul National University, Seoul 08826, Korea*

We present a two-layer fully connected neuromorphic system based on a thin-film transistor (TFT)-type NOR flash memory array with multiple postsynaptic (POST) neurons. Unsupervised online learning by spike-timing-dependent plasticity (STDP) on the binary MNIST handwritten datasets is implemented, and its recognition result is determined by measuring firing rate of POST neurons. Using a proposed learning scheme, we investigate the impact of the number of POST neurons in terms of recognition rate. In this neuromorphic system, lateral inhibition function and homeostatic property are exploited for competitive learning of multiple POST neurons. The simulation results demonstrate unsupervised online learning of the full black-and-white MNIST handwritten digits by STDP, which indicates the performance of pattern recognition and classification without preprocessing of input patterns.

**1. INTRODUCTION**

In the era of exponential data growth, bio-inspired neuromorphic computing system has been suggested as one of the most promising computing architectures for achieving low-power operation.[1-6] With the help of software, neuromorphic systems based on deep neural networks (DNNs) using the back-propagation algorithm have been highlighted for its excellent computational capability.[7-9] The computing system using offline supervised method is suitable for dealing with labeled data, which requires extrinsic error calculation for adjusting parameters before the process.[10] On the other hand, in spiking neural networks (SNNs), complex cognitive tasks involving unsupervised online learning and recognition of unstructured data can be effectively performed using STDP learning algorithm associated with connections between neurons.[11-15] Currently, the inference performance of neuromorphic systems based on SNNs has been investigated in simulations.[16-19] However, these works require additional circuitry for fine-tuning of device parameters given the variability of memristors, which has remarkable impact on the recognition performance for processing various types of data. Previously, we presented a small-scale pattern classification task by introducing a TFT-type NOR flash memory cell as a synaptic device.[20] In this study, a two-layer fully connected neuromorphic system with multi-neuron of an output layer is proposed using the characteristics of the TFT-type NOR flash synaptic devices. In order to show competitive performance of unsupervised online learning by STDP learning algorithm, an update pulse scheme is modified to induce multi-level synaptic weight states and improve linearity of conductance response. Furthermore, homeostatic property of multiple POST neurons has been exploited to classify the full binary MNIST dataset in an unsupervised manner.[16]

---

[*]Corresponding author. Tel: +82-2-880-1727, fax: +82-2-882-4658, e-mail address: jhl@snu.ac.kr

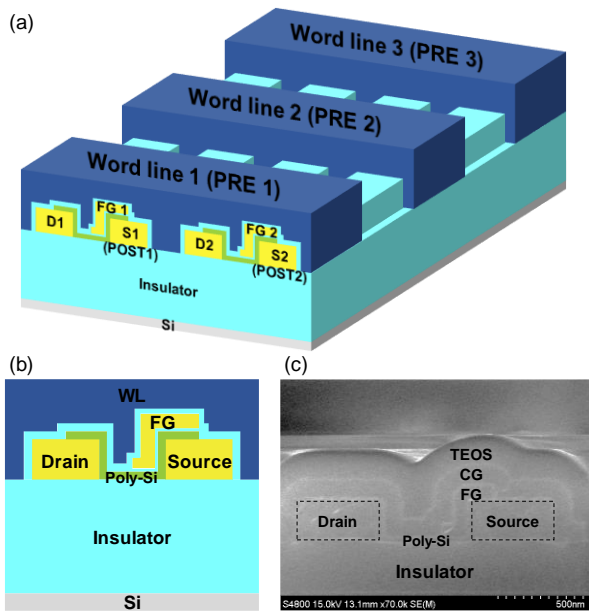

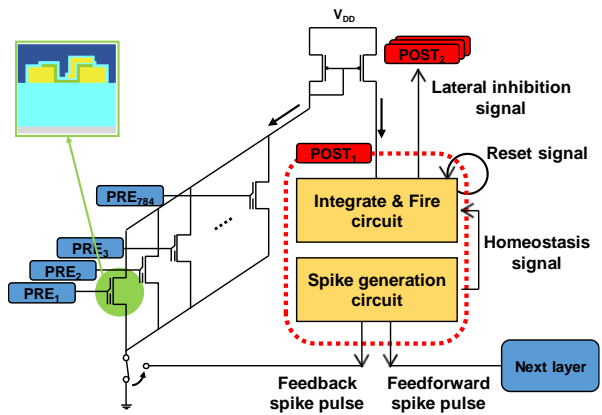

Fig. 1. (a) Schematic 3D array view, (b) cross-sectional single cell view, and (c) SEM cross-sectional image of a TFT-type NOR flash memory synaptic device.

## 2. SPIKING NEURAL NETWORK USING NOR FLASH MEMORY

Fig. 1(a) and (b) show schematic 3D array view and cross-sectional single cell view of a TFT-type NOR flash memory synaptic device, respectively. The crossbar synaptic array in which word line (WL) and bit line (BL) represent the control gate and the drain, respectively, is advantageous for scaling of synaptic devices. As shown in Fig. 1(b), a half-covered $n^+$ poly-Si floating gate (FG) between a cross-point of the WL and the source line serves as a charge storage layer. The amount and polarity of the charge stored can be controlled by adjusting bias conditions of the WL and the source line. Fig. 1(c) shows the fabricated TFT-type NOR flash memory synaptic device in our previous work.[20] The synapse-like nature of the crossbar NOR flash memory arrays is due to the fact that input pulses applied to each PRE neuron reflect the memory state of each cell as a current and are summed up at the BL.

Fig. 2. Schematic diagram of a two-layer unsupervised neuromorphic system with a TFT-type NOR flash memory arrays and a neuron circuit

Fig. 2 introduces the whole neuromorphic system using a TFT-type NOR flash memory synaptic array. In order to implement computational tasks by STDP learning rule, the system is largely divided into three components: synaptic arrays, integrate-and-fire POST neuron circuits, and spike generators. As PRE input pulses are applied into WLs, the current through each synapse is fed into the POST neurons via a current mirror circuit. When the membrane potential exceeds a threshold of the POST neuron, a pulse from the spike generator based on the neuronal firing is transmitted to other neurons. The connectivity of POST neurons leads to competitive learning with lateral inhibition in the network. Homeostatic property is also implemented for balancing neural activity of POST neurons in the output layer. Concurrently, by presenting a feedback pulse of the POST neuron to the source line of the synaptic array, a synaptic weight update based on STDP learning algorithm can be implemented in dependence on the timing difference between input and output spikes. Fig. 3 (a) represents a pulse scheme of PRE and POST neurons for selective updating in accordance with overlapping voltage between PRE ($X_{pre} = V_G$) and POST ($X_{post} = V_S$) neurons. As a POST neuron is fired, the weights of synapses contributing to the spike are potentiated in erase condition ($X_{pre} = -3$ V, $X_{post} = V_{post,+}$) by applying a POST feedback pulse to the common source line, which corresponds to the so-called long-term potentiation (LTP). The weights of the others are depressed in program condition ($X_{pre} = 0$ V, $X_{post} = V_{post,-}$) by the POST feedback pulse, which corresponds to the so-called long-term depression (LTD) in the STDP learning rule. Accordingly, the simplified STDP characteristic is illustrated as shown in Fig. 3(b). The weight change of LTP and LTD are varied with the conductance states of synapses.

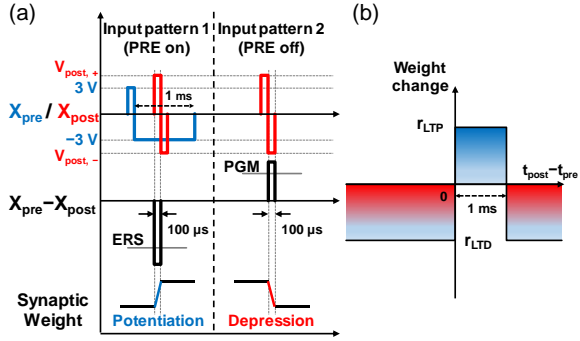

Fig. 3. (a) Pulse scheme of PRE (feedforward input) and POST (feedback output) neurons for STDP learning rule. (b) Simplified STDP curve.

Fig. 4(a) shows measured LTP/LTD characteristics of a synaptic device. Specifically, as shown in Fig. 4(b), the LTD characteristic was investigated in three cases to avoid an abrupt LTD characteristic, which degrades accuracy in our previous work.[20] By modulating the amplitude of negative voltage of feedback POST pulse in case 3 of Fig. 4(b), not only multilevel conductance states were achieved, especially in high level range of weights, but also dynamic range was retained. The conductance behaviors using the condition of the case 3, which indicates high linearity compared with other cases, were maintained for three times of cycles consisting of 50 repeated potentiation pulses ($X_{pre} = -3$ V, $X_{post} = 5$ V) followed by 300 repeated depression pulses ($X_{pre} = 0$ V, $X_{post} = -4.8$ V).

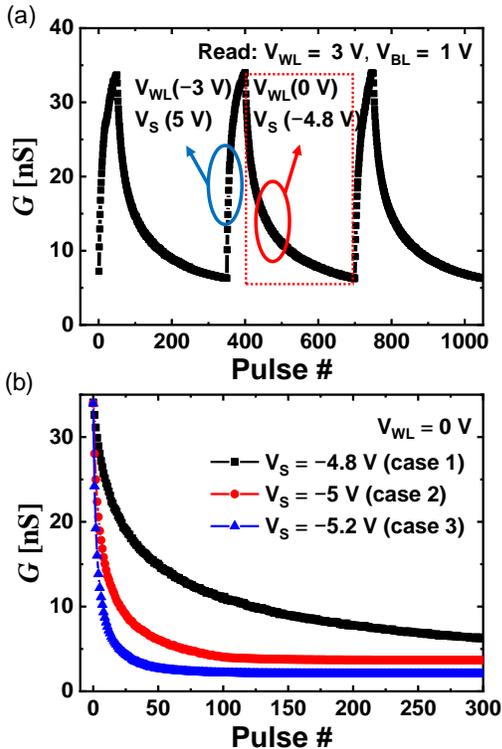

Fig. 4. (a) Measured LTP/LTD characteristics using proposed pulse scheme. (b) LTD characteristics using three different pulse schemes.

### 3. UNSUPERVISED ONLINE LEARNING

To evaluate cognitive performance in the SNN-based system, unsupervised online learning with multiple POST neurons by STDP on full binary MNIST datasets was investigated. The system level pattern learning simulation was performed by the software MATLAB using the measured characteristics of a TFT-type NOR flash memory cell in Fig. 4(a). A fully connected two-layer neuromorphic system based on SNN was designed as in Fig. 5(a). The overall training and recognition processes used in simulations are introduced as illustrated in Fig. 5(b). When a POST neuron fires during the learning, the membrane potentials of all POST are reset by a lateral inhibition mechanism, preventing firing of other POST neurons. In order to ensure balanced use of output neurons, homeostatic property is proposed. Each POST neuron has an exponentially decaying threshold voltage, which is increased as POST neurons fire.[16] In this simulation, 60,000 training examples of full binary MNIST datasets are presented during the training in a feed-forward manner. At the end stage of the learning process, a digit class of a POST neuron is determined by a highly responsive pattern in each POST neuron for the last 10,000 input examples. In the recognition process, all steps of the learning process are excluded except for integrate-and-fire process. 10,000 test examples are used for measuring firing rate of POST neurons and testing how the predicted class of patterns matches the most responsive pattern in the recognition step.

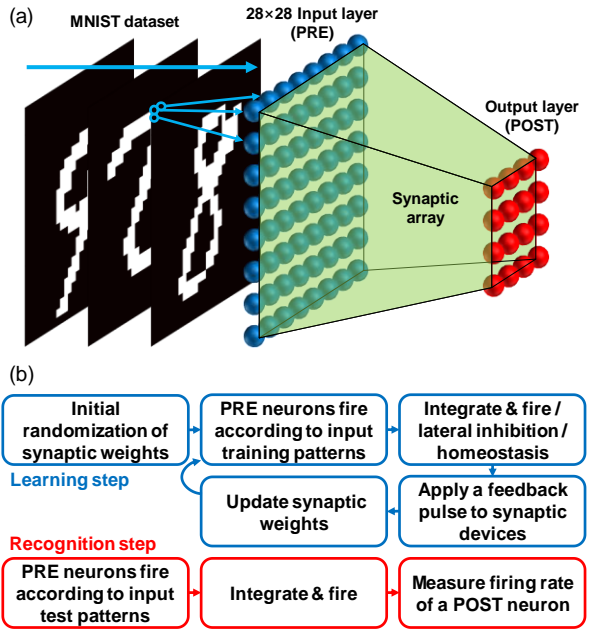

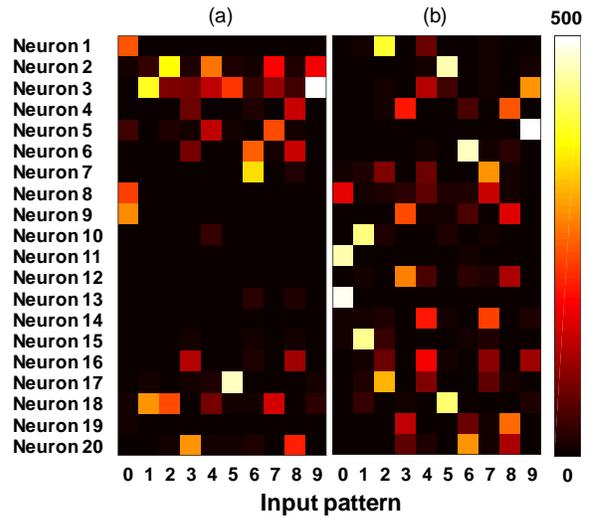

Fig. 5. (a) Schematic illustration of a fully connected two-layer system with a 28×28 PRE layer and a multi-neuron output layer. (b) Flowchart of pattern learning and recognition process.

Fig. 6. Simulation results of unsupervised online learning using 20 POST neurons (a) without and (b) with homeostatic property.

Fig. 6 shows simulation results of unsupervised online learning on MNIST data using 20 POST neurons with homeostatic property. Without the homeostasis, some POST neurons dominate the responses to test patterns and prevent other output neurons from reacting to other input patterns by the lateral inhibition of reactive neurons, which accordingly deactivates some other POST neurons. Most neurons are available in the learning process with the help of the homeostatic property, which leads to improvement of classification in neuromorphic systems using multiple POST neurons. Although this approach requires careful choice of the target POST firing rate, pattern classification with the adaptive threshold improves in comparison to the case without homeostasis. Fig. 7 shows average confusion matrix over 10 digit patterns of the MNIST test set in three different cases mentioned in Fig. 4(b). A smooth depression behavior in case 3 is advantageous compared to an abrupt behavior of other cases. During learning process, when a POST neuron is fired by a pattern input different from the previously learned pattern, the more abrupt the weight update behavior, the easier it is to lose the previous information. In addition, it can be inferred that multi-level states lead to precise learning performance.[9, 21] Fig. 8(a) shows that the larger the output neuron layer, the more types of MNIST patterns can be distinguished, which leads to improvement of the classification performance. As the size of output layer increases, a variation of the recognition rate according to randomly distributed data also decreases. The confusion matrix was also investigated as shown in Fig. 8(b). It is clearly seen that correct classification is achieved by using multi-neuron POST layer whereas some recognition results indicate misclassified examples such as 4 and 9, and 3 and 5, which can be confused to distinguish by their similarity. Using the proposed neuromorphic system and simple learning mechanism, recognition performance of unsupervised learning algorithm was achieved up to 82% for 100 POST neurons, but there is room for improvement, e.g., increasing the number of neuron layers and optimizing the size of each neuron layer in deep multi-layer systems.

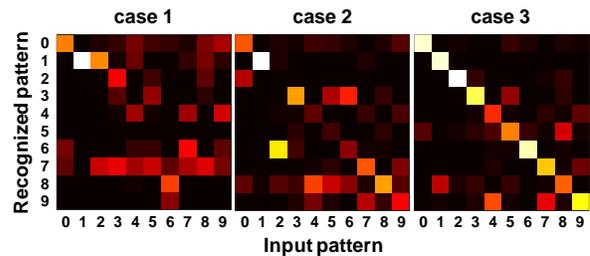

Fig. 7. Confusion matrix of pattern classification results based on three different cases of pulse schemes (Fig. 4(b)) using 20 POST neurons.

## 4. CONCLUSION

In this paper, we have proposed biologically plausible SNN-based neuromorphic system using a TFT-type NOR flash memory synaptic array with multiple POST neurons. To improve the linearity in weight update, a pulse scheme was investigated and modulated by adjusting program pulse for high classification accuracy of SNN-based system. Unsupervised online learning on a full binary MNIST datasets based on STDP was successfully performed by the system level simulation. Lateral inhibition and homeostasis functions were exploited for competitive learning and uniform response of multiple POST neurons. We have investigated the impact of the number of POST neurons in terms of recognition rate. These results support a TFT-type NOR flash memory array as a real-time hardware for unsupervised data classification.

**Acknowledgment:** This research was supported by the MOTIE (Ministry of Trade, Industry & Energy) (10080583) and KSRC (Korea Semiconductor Research Consortium) support program for the development of the future semiconductor device, and the Brain Korea 21 Plus Project in 2018.

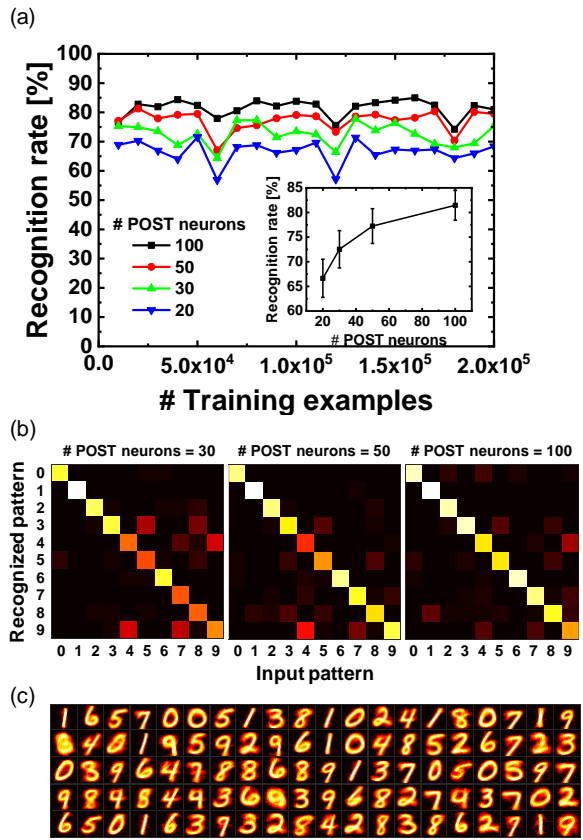

Fig. 8. (a) Recognition rates of proposed SNN for unsupervised online learning. (b) Confusion matrix of the full binary MNIST pattern classification results using a multi-neuron output layer. (c) Synaptic weights after unsupervised online learning with 100 POST neurons.